%% file: main.tex
\documentclass[10pt,twocolumn,letterpaper]{article}

\usepackage{cvpr}              
\usepackage{gensymb}
\usepackage{rotating}

\input{preamble}

\usepackage[pagebackref,breaklinks,colorlinks]{hyperref}

\usepackage[capitalize]{cleveref} 
\crefname{section}{Sec.}{Secs.}
\Crefname{section}{Section}{Sections}
\Crefname{table}{Table}{Tables}
\crefname{table}{Tab.}{Tabs.}
\Crefname{equation}{Equation}{Equations}
\crefname{equation}{eq.}{eqs.}
\crefformat{equation}{Eq.~#2#1#3} 

\newcommand{\green}[1]{\textcolor{ForestGreen}{#1}}
\newcommand{\redtext}[1]{\textcolor{red}{#1}}

\input{dc_macros}

\def\paperID{198} 
\def\confName{3DV\xspace}
\def\confYear{2024\xspace}

\title{SCENES: Subpixel Correspondence Estimation With Epipolar Supervision}

\author{Dominik A. Kloepfer\\
Visual Geometry Group\\
University of Oxford\\
{\tt\small dominik@robots.ox.ac.uk}
\and
Jo\~ao F. Henriques\\
Visual Geometry Group\\
University of Oxford\\
{\tt\small joao@robots.ox.ac.uk}
\and
Dylan Campbell\\
School of Computing\\
Australian National University\\
{\tt\small dylan.campbell@anu.edu.au}
}

\begin{document}
\maketitle
\input{sec/0_abstract}    
\input{sec/1_intro}
\input{sec/2_relatedwork}
\input{sec/3_method}
\input{sec/4_experiments}
\input{sec/5_conclusion}
{
    \small
    \bibliographystyle{ieeenat_fullname}
    \bibliography{main}
}
\input{sec/X_suppl}

\end{document}

%% file: preamble.tex
\usepackage[dvipsnames]{xcolor}

\usepackage{multirow}
\usepackage{tabularx}
\newcolumntype{C}{>{\centering\arraybackslash}X}

\usepackage{tikz}
\usetikzlibrary{quotes, arrows.meta, angles, calc, shapes, intersections, positioning, backgrounds, matrix}
\usepackage{pgfplots}

%% file: sec/0_abstract.tex
\begin{abstract}

Extracting point correspondences from two or more views of a scene is a fundamental computer vision problem with particular importance for relative camera pose estimation and structure-from-motion.
Existing local feature matching approaches, trained with correspondence supervision on large-scale datasets, obtain highly-accurate matches on the test sets.
However, they do not generalise well to new datasets with different characteristics to those they were trained on, unlike classic feature extractors.
Instead, they require finetuning, which assumes that ground-truth correspondences or ground-truth camera poses and 3D structure are available.
We relax this assumption by removing the requirement of 3D structure, \eg, depth maps or point clouds, and only require camera pose information, which can be obtained from odometry.
We do so by replacing correspondence losses with epipolar losses, which encourage putative matches to lie on the associated epipolar line.
While weaker than correspondence supervision, we observe that this cue is sufficient for finetuning existing models on new data.
We then further relax the assumption of known camera poses by using pose estimates in a novel bootstrapping approach.
We evaluate on highly challenging datasets, including an indoor drone dataset and an outdoor smartphone camera dataset, and obtain state-of-the-art results without strong supervision.

\end{abstract}

%% file: sec/1_intro.tex
\section{Introduction}
\label{sec:intro}

\begin{figure}[!t]
  \centering
  \begin{subfigure}[b]{0.45\columnwidth}
    \includegraphics[width=\linewidth]{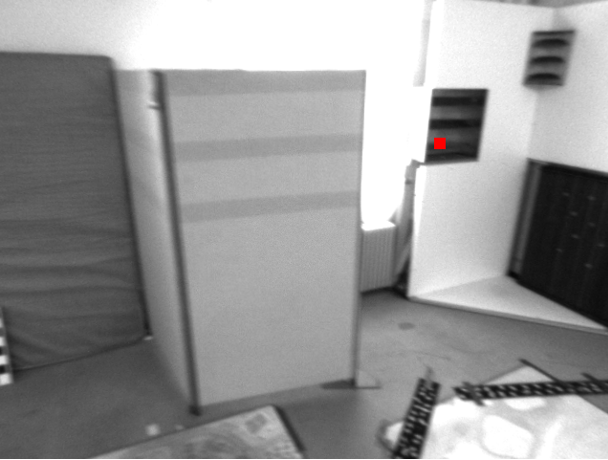}
    \caption{Source pixel}
    \label{fig:subfigure_a}
  \end{subfigure}
  \begin{subfigure}[b]{0.45\columnwidth}
    \includegraphics[width=\linewidth]{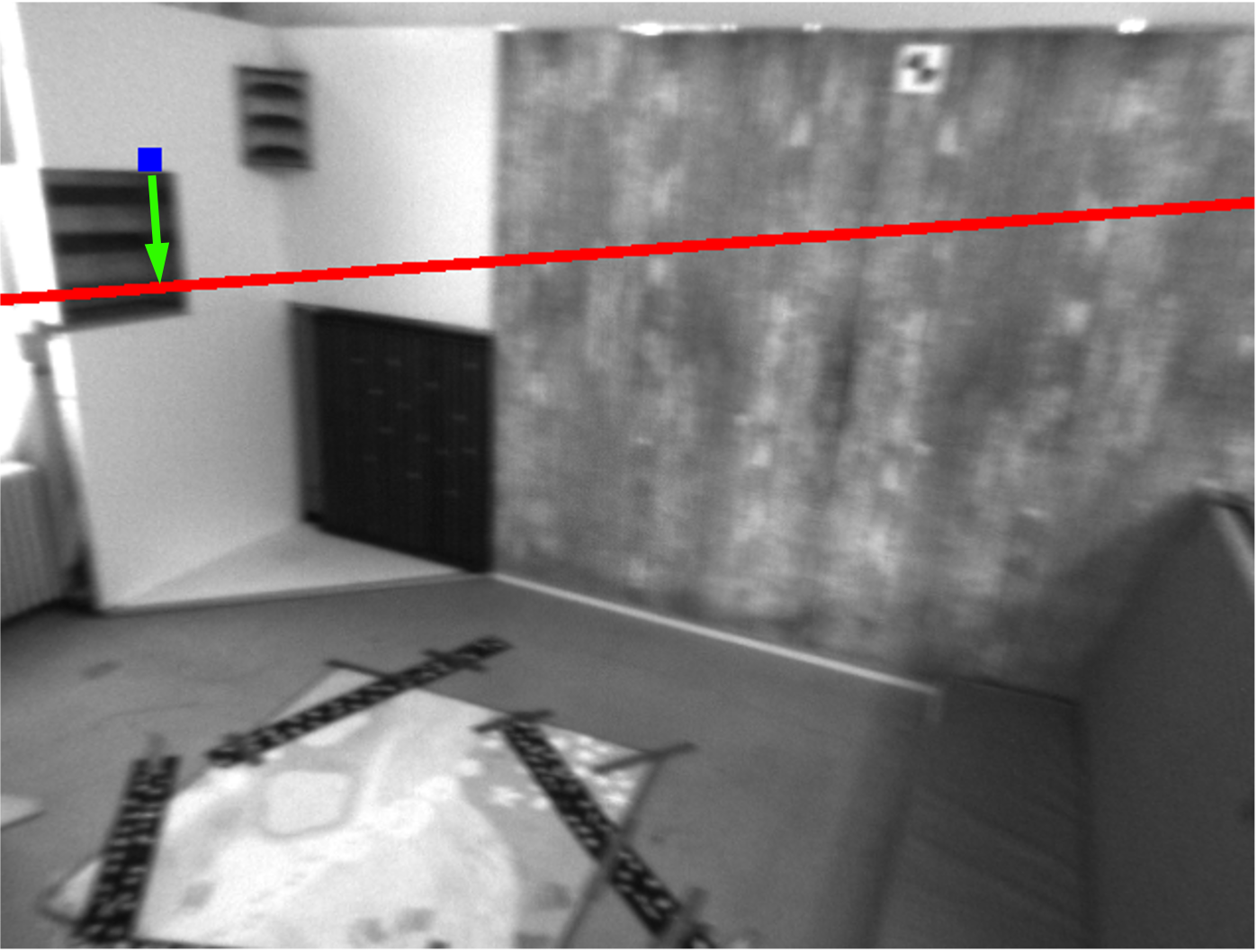}
    \caption{Putative match, epipolar line}
    \label{fig:subfigure_b}
  \end{subfigure}
  \begin{subfigure}[c]{0.9\columnwidth}
    \includegraphics[width=\linewidth]{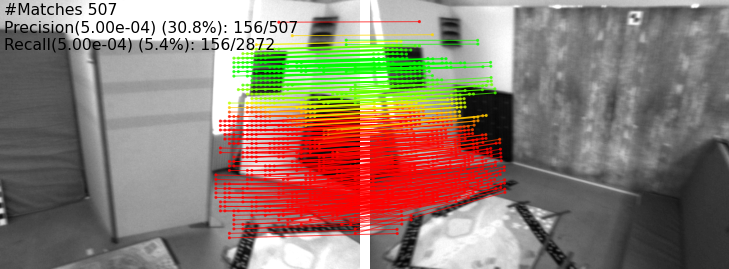}
    \caption{Matches before SCENES finetuning}
    \label{fig:subfigure_c}
  \end{subfigure}
  \begin{subfigure}[d]{0.9\columnwidth}
    \includegraphics[width=\linewidth]{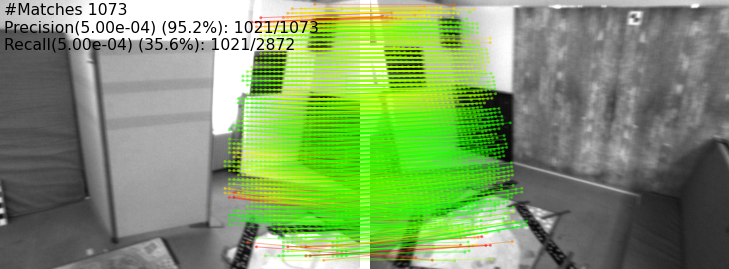}
    \caption{Matches after SCENES finetuning}
    \label{fig:subfigure_d}
  \end{subfigure}
  \caption{SCENES (Subpixel Correspondence EstimatioN with Epipolar Supervision) learns to find high-quality local image matches without requiring correspondence supervision. Instead, pose supervision alone is used, encouraging matches to lie on epipolar lines. The red pixel in (a) corresponds to the red epipolar line in (b). The network initially matches the red pixel to the blue pixel, but our epipolar losses preference matches on the epipolar line (not necessarily the closest point).
  The correspondences found by the state-of-the-art MatchFormer algorithm \cite{wang2022matchformer} are displayed before (c) and after (d) finetuning with epipolar supervision.
  The colour denotes whether the squared symmetrical epipolar distance is below (green) or above (red) a threshold of $5\cdot10^{-4}$.
  Images are from the challenging EuRoC-MAV drone dataset \cite{burri2016eurocmav}.
  }
  \label{fig:splash}
\end{figure}

The task of finding corresponding points in two or more views of a scene---that is, the subpixel positions that correspond to the same visible 3D surface location---is a fundamental low-level image processing problem.
It underpins many computer vision tasks, including relative pose or fundamental matrix estimation \cite{hartley2004multiple,zhang2022relpose,lin2023relposepp}, visual localisation \cite{taira2018inloc,sarlin2019coarse}, Structure-from-Motion (SfM) \cite{schonberger2016structure,sarlin2023pixel}, and Simultaneous Localisation and Mapping (SLAM) \cite{mur2015orb,engel2017direct}, and is related to optical flow estimation \cite{horn1981determining,baker2004lucas,teed2020raft,jiang2021learning} when the camera motion is small.

Classically approached in two stages with keypoint detection followed by local feature extraction and matching \cite{lowe1999object, rublee2011orb, dusmanu2019d2}, recent approaches \cite{sun2021loftr,wang2022matchformer,chen2022aspanformer} eschew the detection step to ensure that dense matches can be extracted even in texture-poor regions.
However, the supervision requirements of most approaches are onerous: ground-truth correspondences are assumed to be available.
While these can be obtained from camera poses and 3D structure (depth maps, point clouds or meshes), these are often not readily accessible.
This makes it challenging to train or finetune these models on new datasets or in new regimes.
For example, a robot preloaded with the latest local feature matching network would not be able to adapt the model when deployed in a new environment, unless it were able to reconstruct the environment.
Obtaining a good quality reconstruction would itself require a good matching network, unless auxiliary sources of information, like a 3D sensor, were available.
To resolve this chicken-and-egg problem, we instead propose an approach that only requires approximate known camera poses, which can be acquired from odometry or GPS combined with an accelerometer.
This allows, in particular, for model adaptation and finetuning in domains where ground-truth 3D structure is difficult to procure.

To address this problem, we propose to fine-tune existing models using epipolar losses, drop-in replacements for the correspondence losses in state-of-the-art transformer-based matching networks.
These encourage putative matches to lie on the associated epipolar lines, which are computed using the relative camera poses.
While significantly weaker than correspondence supervision, we observe that this cue is sufficient for finetuning existing models on new data.
Finally, we propose a bootstrapping approach that removes the need for camera pose supervision as well, allowing us to use an existing pre-trained model to improve itself.
This expands the types of settings where these models can be deployed and reduces the annotation and pre-processing burdens.
To summarise, our contributions are:
\begin{enumerate}
    \item Epipolar losses---drop-in replacements for existing matching losses that remove the need for strong correspondence supervision; and
    \item Strategies for adapting pre-trained models to new domains with only pose supervision, or no supervision.

\end{enumerate}
We evaluate on two challenging datasets with significantly different characteristics from the pre-training data: an indoor drone dataset and an outdoor phone camera dataset.
Compared to the original model, our weakly supervised and unsupervised finetuning strategies allow us to find keypoint matches that achieve significantly improved pose estimation results.

%% file: sec/2_relatedwork.tex
\section{Related work}
\label{sec:relatedwork}

\paragraph{Epipolar geometry in deep learning.}

Several studies have leveraged geometric priors based on epipolar geometry to enhance the performance of deep learning methods across various tasks.
For instance, \citet{he2020epipolartransformers} conduct direct feature aggregation along epipolar lines to improve 3D human pose estimation, while \citet{bhalgat2023light} guide a transformer model to prioritise attending to features along the epipolar line, leading to enhanced 3D object pose estimation.
\citet{yifan2022input} introduce geometrical embeddings based on epipolar geometry in place of positional embeddings for 3D reconstruction tasks using transformer models.
Furthermore, \citet{prasad2018epipolar}, \citet{li2021revisiting}, and \citet{yang2022mvs2d} all employ epipolar geometry for depth prediction tasks.

We also make use of epipolar geometry in our proposed pixel correspondence estimator.
However, we do not explicitly constrain the model to predict matches on the epipolar lines, but rather encourage this behaviour through our epipolar loss functions.
In this respect, our work is most similar to CAPS \citep{wang2020CAPS}, which also learns keypoint feature descriptors.
That approach however is designed for training from scratch, so cannot benefit from pre-training and always requires camera pose supervision, unlike our proposed unsupervised strategy.

\paragraph{Keypoint detection and feature matching.}
The problem of creating correspondences between points in image pairs has been a longstanding challenge in computer vision. Early efforts involved hand-crafted criteria for keypoint detection, description, and matching in well-established methods such as the Harris corner detector~\cite{harris1988combined}, SIFT~\cite{lowe1999object}, SURF~\cite{bay2006surf}, and ORB~\cite{rublee2011orb}.
More recently, deep neural network methods have outperformed classical techniques for both keypoint detection and matching. Approaches like SuperPoint~\cite{detone2018superpoint}, D2Net~\cite{dusmanu2019d2net}, R2D2~\cite{revaud2019r2d2}, and DISK~\cite{tyszkiewicz2020disk} excel at keypoint detection, while works such as SuperGlue~\cite{sarlin2020superglue}, SeededGNN~\cite{chen2021seededGNN}, ClusterGNN~\cite{shi2022clustergnn}, and the very recent LightGlue\cite{lindenberger2023lightglue} achieve impressive feature matching results.
An alternative class of approaches eschew the detection step entirely. These detector-free methods, including LoFTR~\cite{sun2021loftr}, MatchFormer~\cite{wang2022matchformer} and ASpanFormer~\cite{chen2022aspanformer}, have demonstrated remarkable performance by creating matches on a dense grid, eliminating the need for separately selected keypoints, and achieve state-of-the-art accuracy.
Due to their success, we focus on this detector-free paradigm as well, although our approach is not limited to these models.

\paragraph{Unsupervised domain adaptation.}

Finetuning a large model pre-trained on large amounts of data (a computationally expensive process) in order to improve performance or decrease training time is a well-established practice in the field of deep learning, and a comprehensive review would exceed the scope of this paper.
Of particular relevance to the present work is the sub-field of unsupervised domain adaptation (see, \eg, \citet{wilson2020survey1} and \citet{farahani2021survey2} for surveys). In unsupervised domain adaptation, a base model trained on labeled data is fine-tuned to improve performance on a different dataset with limited or no labels. However, the majority of papers in this area focus on methods to make the base model more adaptable to domain shifts, with comparatively less emphasis on the fine-tuning process itself. 
We instead consider how to fine-tune a model that was trained with strong supervision in domains where such ground-truth information is unavailable.

%% file: sec/3_method.tex
\section{Epipolar supervision for matching}
\label{sec:method}

In this section, we present Subpixel Correspondence EstimatioN with Epipolar Supervision (SCENES), our method for estimating image point matches without correspondence supervision.
We first define the problem, then review the relevant background material on epipolar geometry and detector-free feature matching methods.
Next, we present our epipolar loss functions, drop-in replacements for existing matching losses.
Finally, we describe two training settings for adapting pre-trained models to new domains, one with pose supervision and the other without supervision.

The problem under consideration is that of finding the corresponding points in two or more views of a scene: the (subpixel) image positions that correspond to the same visible 3D point.
Without loss of generality, we consider the two image case.
Formally, given two images $\mI_1$ and $\mI_2$ with dimensions $H_1 \times W_1$ and $H_2 \times W_2$ and camera matrices $\mP_1$ and $\mP_2$, we aim to find pairs of (homogeneous) image coordinates $(\bx_1 \in \mI_1, \bx_2 \in \mI_2)$ with high recall and precision such that $\bx_1 = \mP_1 \bX$ and $\bx_2 = \mP_2 \bX$, where $X \in \reals^3$ is a visible surface point on the scene geometry.

\subsection{Background}\label{subsec:background}

\paragraph{Epipolar geometry.}
We begin with a brief review of epipolar geometry; for a more detailed treatment, refer to \citet{hartley2004multiple}.
The epipolar geometry of two views is the geometry of the intersection of the two image planes with the pencil of planes through the baseline joining the camera centres.
For a 3D point $\bX$ imaged in two views at 2D points $\bx_1$ and $\bx_2$, the epipolar geometry defines the relation between these image points via the fundamental matrix $\mF$.
Expressing $\bx_1$ in homogeneous coordinates and interpreting it as a ray extending from the camera centre through $\bx_1$ in the image plane, then the projection of this line into a second image is the epipolar line corresponding to $\bx_1$.
The corresponding point $\bx_2$ must lie on this line, a relation that is captured by the equation
\begin{equation}
\label{eq:epipolar_relation}
\bx_{2}\transpose \mF \bx_1 = 0
\end{equation}
where the epipolar line corresponding to the point $\bx_1$ in image 2 is given by
$\bl_{12} = \mF \bx_1$.
This constitutes a significant constraint on where the corresponding point might be located, reducing the search space from 2D to 1D.

The fundamental matrix relating two images can be computed directly from the images, either from (at least seven) point correspondences or from their relative pose and camera intrinsics, without requiring any information about the 3D geometry of the scene.
That is, given two camera matrices $\mP_1 = \mK_1 [\mI \mid \zeros]$ and $\mP_2 = \mK_2 [\mR \mid \bt]$ with a rotation $\mR$, translation $\bt$ (up to scale) and camera intrinsic matrices $\mK_1$ and $\mK_2$, the fundamental matrix $\mF \in \reals^{3\times 3}$ is given by
\begin{equation}
\label{eq:fmat}
\mF = \mK_{2}^{-\mathsf{T}} [\bt]_\times \mR \mK_{1}^{-1},
\end{equation}
with $[\bt]_\times = \left[ \begin{smallmatrix} 0 & -t_3 & t_2 \\ t_3 & 0 & -t_1 \\ -t_2 & t_1 & 0 \end{smallmatrix} \right]$ is the matrix representation of the cross product.
Since a good estimate of the relative camera pose is often available from odometry and the intrinsic parameters are often recorded in the image metadata, then the fundamental matrix and hence the epipolar lines can be easily obtained.

\paragraph{Local feature matching.}


State-of-the-art detector-free image matching algorithms, starting with LoFTR \cite{sun2021loftr} and including MatchFormer \cite{wang2022matchformer} and ASpanFormer \cite{chen2022aspanformer}, aim to find as many accurate matches as possible for $m$ pixels in the first image, arranged in a grid at locations $\bx_1^i \in \mI_1$ with $i=1,\dots, m$.
They involve two stages: coarse and fine.

In the first stage, these methods estimate a set of high-level matches by extracting coarse features, generally for $8\times8$ pixel patches.
These coarse features are then used to construct a confidence matrix $\mC \in \reals^{m \times m}$ that for each sampling location $\bx_1^i$ in the first image estimates the probability that the respective pixel patch at location $\bx_2^j \in \mI_2$ in the second image is a correct match.
This coarse-level confidence matrix is supervised using a classification-type loss $\mathcal{L}_c$, such as cross-entropy or negative log-likelihood, with respect to a ground-truth binary mask $\mM \in \reals^{m \times m}$ that is $1$ for ground-truth correspondences (calculated using ground-truth camera poses and depths) and $0$ otherwise.
Letting $\bx_g^i \in \mI_2$ be the location of the ground-truth match for location $\bx_1^i$, we have
$
\mM_{ij} = 1 \textrm{ iff } \bx_2^j = \bx_g^i.
$
The coarse level loss is then a function of this ground-truth matrix and the confidence matrix:
$
\cL_c = f(\mC, \mM).
$

In the second stage, an initial set of $m' \leq m$ coarse matches computed from the coarse-level confidence matrix are refined at a finer level.
From a set of finer features (generally $2 \times 2$ or $4 \times4$ pixel patches), the second stage crops windows surrounding each of the coarse level matches, and returns a refined set of matched image coordinates $\hat{\bx}_2^k$ with $k=1,\dots, m'$.
Whereas the coarse stage is only designed to compute matches on a discrete grid, the fine stage returns floating-point coordinates.
The supervision of this stage therefore also differs from that of the coarse stage: instead of a classification-type loss, a regression loss $\cL_f$ is used, which is a function of the distance $d$ between the location of the ground-truth match and the refined matching coordinates:
$
\cL_f = \frac{1}{m'} \sum_k^{m'} g(d(\hat{\bx}_2^k, \bx_g^k)).
$

\subsection{Epipolar losses}\label{subsec:epipolar_losses}

\begin{figure}[!t]\centering%
\begin{subfigure}{0.33\linewidth}
\input{figures/epipolar_loss_classification_a.tikz}\unskip%
\caption{Classification: $\mC$}%
\label{fig:losses-a}
\end{subfigure}\hfill
\begin{subfigure}{0.33\linewidth}
\input{figures/epipolar_loss_classification_b.tikz}\unskip%
\caption{Classification: $\mM^\textrm{epi}$}%
\label{fig:losses-b}
\end{subfigure}\hfill
\begin{subfigure}{0.33\linewidth}
\input{figures/epipolar_loss_regression.tikz}\unskip%
\caption{Regression: $d^\textrm{epi}$}%
\label{fig:losses-c}
\end{subfigure}
\caption{
Visualisation of the epipolar classification (coarse) and regression (fine) losses.
The epipolar classification loss assigns the `ground-truth' match location to the highest probability point on the epipolar line and compares the resulting binary mask $\mM^\textrm{epi}$ to the predicted confidence map $\mC$ via a cross-entropy loss or similar.
The epipolar regression loss computes the perpendicular distance between the predicted match $\hat{\bx}_2$ and the epipolar line $\bl_{12}$.
Note the thin epipolar line is plotted in (a) and (b) for visualisation purposes only, it is not part of the confidence map $\mC$ or binary mask $\mM$.
}%
\label{fig:losses}
\end{figure}
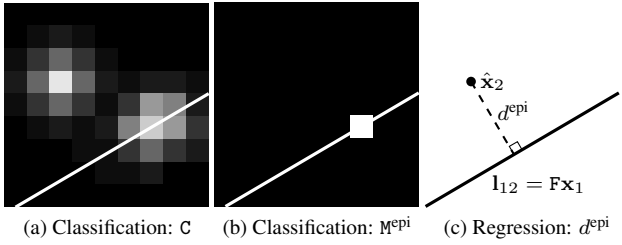

The aforementioned classification and regression losses used to train state-of-the-art local feature matching models require the ground-truth correspondences.
However, unless the dataset has ground-truth camera poses and 3D geometry, these can be difficult to obtain.
To circumvent this limitation, we reformulate both loss types such that they only require the fundamental matrix $\mF$ for a given image pair $\mI_1$ and $\mI_2$, by replacing the ground-truth correspondence matrix $\mM$ and the distance to the ground-truth match $d$ with \textit{epipolar variants} $\mM^\textrm{epi}$ and $d^\textrm{epi}$, as visualised in \cref{fig:losses}. The fundamental matrix $\mF$ of an image pair can be computed from the relative camera pose, or estimated using a keypoint matches between the two images.

\paragraph{Epipolar classification loss.}

For the coarse classification loss, a simple strategy would be to replace the binary mask $\mM$ of the ground-truth point correspondences with a binary mask of the epipolar lines computed using the fundamental matrix.
With $e_1^i \subset \mI_2$ the set of coordinates in the second image that form the epipolar line for point $\bx_1^i \in \mI_1$, this would require $\mM^\textrm{epi}_{ij} = 1 \textrm{ iff } \bx_2^j \in e_1^i$.
However, fine-tuning the model in this way would train it to predict equally high matching scores for every pixel along the epipolar line. This means that the probability mass for the coarse matches is diffused along the entire epipolar line, reducing the reliability of the coarse matches.
Instead, we assign the ground-truth match location at the position on the epipolar line that has the highest probability, taking inspiration from the max-epipolar loss of \citet{bhalgat2023light}.
That is, we define 
\begin{align}
\mM^\textrm{epi}_{ij} =
\begin{cases}
1 & \textrm{if } j =  \argmax_{k:\, \bx_2^k \in e_1^i}(\mC_{ik})\\
0 & \textrm{otherwise.}
\end{cases} \label{eq:epipolar_classification}
\end{align}

\paragraph{Epipolar regression loss.}

For the fine-level regression loss, we replace the distance between the estimated and the ground-truth match with the perpendicular distance between the estimated match $\hat{\bx}_2^i$ and the epipolar line $\bl_{12}^i$, as was also done in \citet{wang2020CAPS}.
In homogeneous coordinates, with the third coordinate normalised to 1, 
\begin{equation}
d^\textrm{epi}(\bx_1^i, \hat{\bx}_2^i) = \frac{|\bx_1^{i\mathsf{T}} \mF^{\mathsf{T}} \hat{\bx}_2^i|}{\sqrt{([\mF^{\mathsf{T}} \hat{\bx}_2^i]_1)^2 + ([\mF^{\mathsf{T}} \hat{\bx}_2^i]_2)^2}}.
\end{equation}

Here $[\mF^{\mathsf{T}} \hat{\bx}_2^i]_k$ refers to the $k$-th component of the three-dimensional vector $[\mF^{\mathsf{T}} \hat{\bx}_2^i]$.

\paragraph{Total epipolar loss.}

The total epipolar loss is then 
\begin{equation}
\cL^\textrm{epi} = (1 - \lambda) f(\mC, \mM^\textrm{epi}) + \frac{\lambda}{m'} \sum_k^{m'} g(d^\textrm{epi}(\bx_1^k, \bx_2^k))
\end{equation}
where typically the function $f$ is a binary cross entropy function, $g$ is a simple linear scaling, and $\lambda$ is a hyper-parameter that controls the relative importance of the classification and regression losses.
We note that it is generally straightforward to adapt most loss functions that use ground-truth image correspondences to supervise keypoint matches.

\subsection{Finetuning with weak supervision}

Reformulating the losses in terms of the epipolar lines and distances lessens the supervision requirements.
This facilitates the adaptation of pre-trained models to new domains, where ground-truth correspondences may be difficult to obtain.
Instead, given a fundamental matrix relating two images, which can be obtained from the relative pose between the cameras (a rotation $\mR$ and translation $\bt$ up to scale) and the camera intrinsic matrices $\mK_1$ and $\mK_2$ as shown in \cref{eq:fmat}, both $\mM^\textrm{epi}$ and $d^\textrm{epi}$ can be computed. 
A pre-trained model can then be finetuned on a new dataset by optimising $\cL^\textrm{epi}$.

\subsection{Finetuning without supervision}\label{subsec:bootstrapping}

In scenarios where accurate ground-truth camera poses are unavailable, the model can still be refined via the epipolar loss using a bootstrapping strategy.
While the ground-truth fundamental matrix $\mF$ cannot be computed without ground-truth camera poses, it can be estimated from (at least seven) correspondences, such as those found by a pre-trained matching model.
Our strategy, then, is to apply a pre-trained model to a dataset from a new domain to obtain a set of (poor quality) image correspondences, then use a robust estimator such as RANSAC to estimate the fundamental matrices, which are used as supervision for our epipolar loss.
The key intuition is that even if the pre-trained model elicits many incorrect matches for an image pair, the robust estimator should still be able to estimate the (approximately) correct fundamental matrix in many cases, which will provide a supervision signal for the incorrect matches.
%
As we show in \cref{sec:experiments}, while this bootstrapping method is naturally less performant than finetuning using ground-truth fundamental matrices, it can still lead to significant performance improvements on challenging datasets.

\subsection{Discussion}
While we do not know the ground-truth locations of the keypoint matches in the absence of ground-truth depth and/or ground-truth pose information, we do know that they must lie on the respective epipolar lines.

For the coarse stage, we want to refine the network's predicted confidence matrix to align with our best estimate of the location of the match.
Since we are starting from a pre-trained network that does not make random predictions, this best estimate is the location with the highest probability for being a match that is also on the epipolar line, \ie, the location on the epipolar line for which the matching confidence $\mC_{ij}$ is maximised.
This provides a useful training signal so long as the highest probability location on the epipolar line is near the correct match location sufficiently frequently.

For the fine stage, replacing the distance to the ground-truth match location with the distance to the epipolar line in the regression loss also provides an imperfect-yet-useful training signal.
Even though the gradient vector when taking the derivative of $d^\textrm{epi}$ \wrt $\bx_2^k$ may not point exactly in the same direction as the gradient vector from taking the derivative of $d$ \wrt $\bx_2^k$, the angle between them is less than ninety degrees (they have a positive inner product $\langle \frac{\partial d}{\partial \bx_2^k}, \frac{\partial d^\textrm{epi}}{\partial \bx_2^k}\rangle > 0$).
Following the gradient vector of the epipolar loss therefore also tends to decrease the distance of the match to the ground-truth match location.

The bootstrapping approach, using estimated instead of ground-truth fundamental matrices, ultimately improves the keypoint matches generated by improving the consistency of the matches.
By design, algorithms such as RANSAC ignore many or most available point-matches and only use a subset to compute the final hypothesis for the fundamental matrix. 
While the points that were used for fundamental matrix estimation will experience minimal loss, the remaining matches receive a rich gradient signal during training.
In an effort to improve the consistency of all generated matches with the subset of matches that was used to estimate the fundamental matrices, the network will learn to adapt to the new environment's idiosyncrasies. 

%% file: figures/epipolar_loss_classification_a.tikz
\begin{tikzpicture}[scale=0.33-0.25/9]
    \foreach \y [count=\n] in {
        {0,0,0,0,0,0,0,0,0},
        {0,5,10,5,0,0,0,0,0},
        {5,20,40,20,5,0,0,0,0},
        {10,40,90,40,10,5,10,10,5},
        {5,20,40,20,5,20,60,50,10},
        {0,5,10,5,10,50,80,60,20},
        {0,0,0,5,5,20,40,20,10},
        {0,0,0,0,5,5,10,0,0},
        {0,0,0,0,0,0,0,0,0},
    }{
        \foreach \x [count=\m] in \y {
            \node[fill=white!\x!black, minimum size=\linewidth/9] at (\m,-\n) {};
        }
    }
    \begin{scope}[very thick, white]
        \draw (1, -9.5) -- (9.5,-4.5);
    \end{scope}
\end{tikzpicture}

%% file: figures/epipolar_loss_classification_b.tikz
\begin{tikzpicture}[scale=0.33-0.25/9]
    \foreach \y [count=\n] in {
        {0,0,0,0,0,0,0,0,0},
        {0,0,0,0,0,0,0,0,0},
        {0,0,0,0,0,0,0,0,0},
        {0,0,0,0,0,0,0,0,0},
        {0,0,0,0,0,0,0,0,0},
        {0,0,0,0,0,0,100,0,0},
        {0,0,0,0,0,0,0,0,0},
        {0,0,0,0,0,0,0,0,0},
        {0,0,0,0,0,0,0,0,0},
    }{
        \foreach \x [count=\m] in \y {
            \node[fill=white!\x!black, minimum size=\linewidth/9] at (\m,-\n) {};
        }
    }
    \begin{scope}[very thick, white]
        \draw (1, -9.5) -- (9.5,-4.5);
    \end{scope}
\end{tikzpicture}

%% file: figures/epipolar_loss_regression.tikz
\begin{tikzpicture}[scale=0.33-0.25/9, square/.style={regular polygon,regular polygon sides=4}]
    \draw [very thick, black] (1, -9.5) -- (9.5,-4.5);
    \fill (3, -4) circle [radius=6pt];
    \draw [thick, dashed, black] (3, -4) -- (4.9,-7.23);
    \node[anchor=west] at (3.75, -5.275) {\footnotesize $d^\textrm{epi}$};
    \node[anchor=west] at (3, -4) {\footnotesize $\hat{\bx}_2$};
    \node[anchor=west] at (3.5, -8.5) {\footnotesize $\bl_{12} = \mF\bx_1$};
    \node[anchor=south west] at (4.9, -7.23) [square, draw, minimum size=5pt, inner sep=0pt, rotate=30, xshift=-0.2pt] () {};
\end{tikzpicture}

%% file: sec/4_experiments.tex
\section{Experiments}
\label{sec:experiments}

To demonstrate the utility of the SCENES fine-tuning for downstream tasks, we compare the performance of a number of keypoint correspondence estimation methods 

In this section, we demonstrate the utility of SCENES fine-tuning by comparing the performance of fine-tuned models on the downstream task of relative camera pose estimation with that of a number of existing keypoint correspondence estimation methods.
We evaluate on three datasets: the EuRoC-MAV drone dataset \cite{burri2016eurocmav}, the San Francisco Landmark dataset \cite{chen2011sanfrancisco}, and the Aachen day--night dataset \cite{sattler2012aachen1,sattler2018aachen2}
in the supplement.
We then conduct an ablation study on the fine-tuning algorithm to assess the design decisions.

\subsection{Pose-only supervision: EuRoC-MAV dataset}\label{sec:exp-euroc-mav}

The EuRoC-MAV drone dataset \cite{burri2016eurocmav} comprises eleven image sequences captured by a remotely-operated drone in two distinct environments: a spacious machine room and a smaller room. Leveraging a laser tracker in the machine room and a motion capture system in the other room, accurate drone (and camera) poses were meticulously recorded.
The machine room significantly differs in scale and content from ScanNet \cite{dai2017scannet}. It is much larger than the offices, apartments, etc. included in ScanNet, and contains industrial machinery that is not commonly found in such locations. 
The image sequences vary in difficulty based on their lighting levels and amount of motion blur. 
Even the images in the `easy' sequences differ significantly from those in the ScanNet dataset, especially with respect to the distribution of camera poses, when comparing drone footage to handheld camera images.

To use the dataset for correspondence matching, we need to generate image pairs that have enough overlap to make matching keypoints possible, but not so much that the task becomes trivial (\ie, nearly-identical images). Without ground-truth depth information, we cannot calculate the true overlap between the images. Instead, we compute a `pseudo-overlap' score using the known camera poses, which visual inspection shows yields a good range of overlaps between image pairs.
More details on the dataset construction can be found in the supplementary material.

\paragraph{Implementation details.}
We fine-tune the MatchFormer-lite method \cite{wang2022matchformer} pre-trained on the indoor ScanNet dataset \cite{dai2017scannet}, and evaluate using the images from two held-out sequences of the highest difficulty, one from the machine room and one from the indoor room.
We leave all training settings from the original training run unchanged, with the exception of decreasing the weight decay from $0.1$ to $0.01$ and decreasing the learning rate from $3\times 10^{-3}$ to $1\times 10^{-4}$. Computational limitations also mean that we reduced the batch size from $64$ to $12$.
We also fine-tune the ASpanFormer method \cite{chen2022aspanformer}, also pretrained on the indoor ScanNet dataset \cite{dai2017scannet}. 
Again, we mostly retain the training settings from the original training run, except that we reduce the batch size to $8$ due to computational limitations.

\paragraph{Baselines.}
We compare with a number of state-of-the-art image matchers on the dataset. 
Since its loss function only requires camera poses, we train the CAPS~\cite{wang2020CAPS} method from scratch on EuRoC-MAV using the authors' settings. For the other baselines, we use the implementations and pre-trained weights (not trained on EuRoC-MAV) provided by the respective authors, choosing the `indoor' versions where there is a choice.
For every method we compute relative camera pose estimates from the matched keypoints using a five-point essential matrix estimator with RANSAC, as implemented in the OpenCV~\cite{opencv_library} library, with a matching threshold of $0.5$ pixels.

\paragraph{Results.}

\begin{table}[!t]
  \centering\small
  \caption{Relative pose estimation performance on the indoor EuRoC-MAV dataset \cite{burri2016eurocmav}. We report the area under the curve (AUC) at thresholds of $5\degree$, $10\degree$ and $20\degree$ and the matching precision (P) at a threshold of $5\cdot 10^{-4}$. Our SCENES method exhibits significant gains in performance across these metrics.}
  \label{tab:results_eurocmav}
  \begin{tabularx}{\linewidth}{lCCCC}
    \toprule
    \multirow{2}{*}{Method} & \multicolumn{3}{c}{Pose Estimation AUC (\%)} & \multirow{2}{*}{P (\%)} \\
    \cmidrule(r){2-4} 
                        & $@5\degree$ & $@10\degree$ & $@20\degree$  &  \\
    \midrule
    SP~\cite{detone2018superpoint} + SG~\cite{sarlin2020superglue}      & 0.6   & 2.0   & 4.4   & 17.3  \\
    SP~\cite{detone2018superpoint} + CAPS~\cite{wang2020CAPS} & 1.4 & 6.4 & 16.3 & 20.6 \\
    SP~\cite{detone2018superpoint} + LG~\cite{lindenberger2023lightglue}      & 7.6   & 22.9  & 40.2  & 38.3  \\
    DISK~\cite{tyszkiewicz2020disk} + LG~\cite{lindenberger2023lightglue}    & 6.6   & 17.9  & 31.6  & 35.9  \\
    MatchFormer-large~\cite{wang2022matchformer}   & 5.1   & 16.0  & 31.8  & 38.4  \\
    \midrule
    MatchFormer-lite~\cite{wang2022matchformer}   & 3.0   & 11.6  & 25.1  & 35.0  \\
    + SCENES fine-tuning    & \green{+6.1}  & \green{+11.9}  & \green{+17.7}  & \green{+28.8}  \\
                    = Ours (MF)        & \underline{9.1}          & \underline{23.5} & \underline{42.8}  & \underline{63.8}  \\
    \midrule
    ASpanFormer~\cite{chen2022aspanformer} & 4.8   & 17.6  & 34.5  & 38.7  \\
    + SCENES fine-tuning   & \green{+6.8} & \green{+8.8} & \green{+10.2} & \green{+18.1} \\
    = Ours (ASF) & \textbf{11.6}  & \textbf{26.2}  & \textbf{44.7}  & \textbf{66.8}  \\ 
    \bottomrule
  \end{tabularx}
\end{table}

\begin{figure}[!t]
  \centering
\begin{subfigure}[a]{0.796\columnwidth}
    \includegraphics[width=\linewidth]{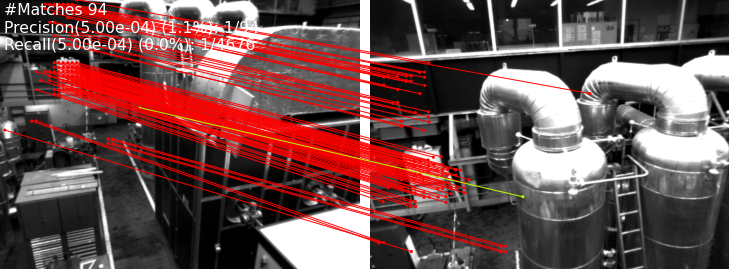}
    \caption{Matches before SCENES fine-tuning}
    \label{fig:qualitative_eurocmav_a}
  \end{subfigure}
  \begin{subfigure}[b]{0.796\columnwidth}
    \includegraphics[width=\linewidth]{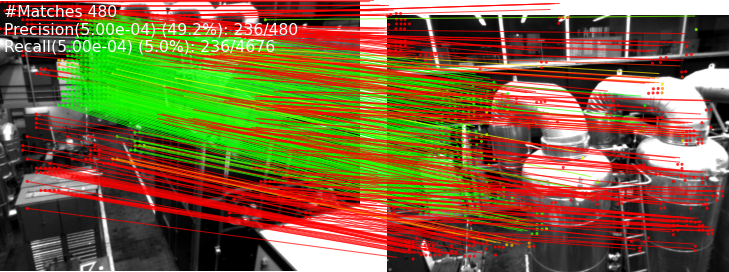}
    \caption{Matches after SCENES fine-tuning}
    \label{fig:qualitative_eurocmav_b}
  \end{subfigure}
\begin{subfigure}[c]{0.796\columnwidth}
    \includegraphics[width=\linewidth]{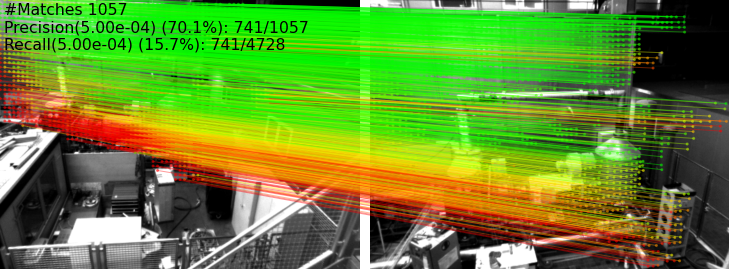}
    \caption{Matches before SCENES fine-tuning}
    \label{fig:qualitative_eurocmav_c}
  \end{subfigure}
  \begin{subfigure}[d]{0.796\columnwidth}
    \includegraphics[width=\linewidth]{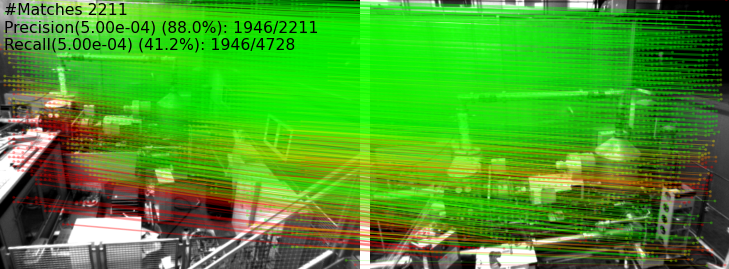}
    \caption{Matches after SCENES fine-tuning}
    \label{fig:qualitative_eurocmav_d}
  \end{subfigure}
  \caption{Qualitative matching results on the EuRoC-MAV drone dataset \cite{burri2016eurocmav}. The correspondences found by the state-of-the-art MatchFormer algorithm \cite{wang2022matchformer} are displayed before (a, c) and after (b, d) SCENES fine-tuning. The colour denotes whether the squared symmetrical epipolar distance is below (green) or above (red) a threshold of $5\cdot10^{-4}$. Our method improves the quality and number of correct matches.
  }
  \label{fig:qualitative_eurocmav}
\end{figure}

Like previous works~\cite{sarlin2020superglue, sun2021loftr, wang2022matchformer, chen2022aspanformer, lindenberger2023lightglue}, we report the area under the cumulative curve (AUC) of the pose error (the maximum of the angular error in rotation and translation) at thresholds of $5\degree$, $10\degree$ and $20\degree$. We also report the matching precision (P): the average percentage of predicted matches with a squared symmetrical epipolar distance smaller than $5\cdot 10^{-4}$.
As can be seen in \cref{tab:results_eurocmav} and in the qualitative results in \cref{fig:qualitative_eurocmav}, the fine-tuned methods show a large jump in performance compared to the original pre-trained methods. The fine-tuned version of the small MatchFormer-lite network even outperforms the previous state-of-the-art, the combination of the SuperPoint detector and the LightGlue matcher.
The performance of the baseline methods on this dataset is considerably worse than on the dataset they were trained on (ScanNet), showing that their ability to generalise to novel data distributions is limited. 
Our SCENES fine-tuning approach goes some way to alleviating this issue.
The CAPS algorithm also struggles, not having benefited from pre-training with ground-truth point correspondences on a different dataset.

\subsection{Bootstrapped supervision: SF dataset}\label{sec:exp-san-fran}

The San Francisco Landmark dataset \cite{chen2011sanfrancisco} is a collection of roughly one million perspective images obtained from panorama images collected by a car in downtown San Francisco. 
The GPS measurements for the camera position and orientation are also provided, though these measurements are frequently inaccurate by up to dozens of metres.
Test images were collected using camera phones, and \citet{torii2019sanfranrefposes} provides accurate reference poses for 598 of them.
While the GPS measurements are too inaccurate to compute fundamental matrices, we can still use them to generate image pairs using a similar `pseudo-overlap' score as before. The reader is referred to the appendix for details.

This dataset represents a realistic example of a setting without any pose supervision available for training or fine-tuning, while still containing a small test dataset with reference poses that allows us to evaluate the effect of the fine-tuning. We give results of our bootstrapping method applied to the EuRoC-MAV dataset (where accurate poses can be collected much more easily) in the supplementary material.

\paragraph{Implementation details.}

We again fine-tune the Matchformer-lite \cite{wang2022matchformer} and ASpanFormer \cite{chen2022aspanformer} methods, this time using the respective `outdoor' weights trained on the MegaDepth dataset \cite{MegaDepthLi18} as initialisation. 
To employ the bootstrapping strategy described in \cref{subsec:bootstrapping} in the absence of ground-truth poses for training images, we use each of the base models to compute fundamental matrices for one million randomly selected image pairs using OpenCV's implementation of RANSAC \cite{fischler1981random}. 
To maximise the reliability of these estimates, we only use those image pairs for training for which the base model extracted at least $100$ keypoint matches and for which the estimated fundamental matrices result in at least $20$ inliers. This leaves $749,581$ and $688,466$ image pairs for fine-tuning the Matchformer-lite and ASpanFormer methods respectively.

The different dataset collection methods for the training and testing datasets lead to meaningfully different data distributions. To ensure that the networks are fine-tuned to adjust to the San Francisco environment without overfitting on features such as the camera location, we add an equal number of image pairs from the MegaDepth dataset \cite{MegaDepthLi18} to each batch. The losses for these MegaDepth images are the original correspondence losses. 

For the Matchformer-lite model, we use the same training settings as during the initial training, with the exception of again decreasing the weight decay to $0.01$ and the learning rate to $1\cdot 10^{-4}$. We also decreased the batch size to $8$ image pairs taken from each of the San Francisco and the MegaDepth datasets.
For the ASpanFormer model, we decreased the batch size further to two image pairs from each of the two datasets, and also decreased the fraction of coarse matches that are used for the fine-level supervision during training from $30\%$ in the original training run to $10\%$.
Due to computational limitations, we also had to change the function $f$ used to compute the coarse level loss from a cross entropy function to the sparse formulation used in LoFTR~\cite{sun2021loftr}.

\paragraph{Baselines.}

We provide results for the same set of state-of-the-art baselines as for the EuRoC-MAV dataset, computing the relative camera poses using RANSAC as before. 
This time, we choose the `outdoor' versions when multiple sets of weights are available.
To aid comparability between different methods, we pad all images to $640\times640$ pixels.

\paragraph{Results.}

We report the area under the cumulative curve (AUC) of the pose error at thresholds of $5\degree$, $10\degree$ and $20\degree$. Following \citet{sarlin2020superglue}, the threshold for the matching precision in this outdoor setting is $1\cdot 10^{-4}$. 

This setting, where the images of the training dataset were collected in a very different way to those in the testing dataset, does not play to the strengths of a fine-tuning approach, as an important type of regularity that the network might adapt to is now the source of misleading training signals.
Despite this, and despite the additional noise resulting from the bootstrapping approach, the results in \cref{tab:results_sanfrancisco} show noticeable performance improvements after fine-tuning. When fine-tuning with bootstrapping on the EuRoC-MAV dataset (see the supplementary), where training and testing sets were collected in more similar ways, the performance gains are even more substantial. 

These results show the feasibility of our bootstrapping strategy for fine-tuning a keypoint matcher on datasets without any ground-truth information.

\begin{table}[!t]
  \centering\small
  \caption{Relative pose estimation performance on the outdoor San Francisco Landmarks dataset \cite{chen2011sanfrancisco}. We report the area under the curve (AUC) at thresholds of $5\degree$, $10\degree$ and $20\degree$ and the matching precision (P) at a threshold of $10^{-4}$. Even without any available ground-truth correspondences, poses or 3D structure, our SCENES fine-tuning method is able to improve the performance of matching algorithms on this challenging dataset.}
  \label{tab:results_sanfrancisco}
  \begin{tabularx}{\linewidth}{lCCCC}
    \toprule
    \multirow{2}{*}{Method} & \multicolumn{3}{c}{Pose Estimation AUC (\%)} & \multirow{2}{*}{P (\%)} \\
    \cmidrule(r){2-4} 
                        & $@5\degree$ & $@10\degree$ & $@20\degree$  &  \\
    \midrule
    SP~\cite{detone2018superpoint} + SG~\cite{sarlin2020superglue}      & 1.2   & 4.1   & 9.9   & 15.1  \\
    SP~\cite{detone2018superpoint} + LG~\cite{lindenberger2023lightglue}      & 19.4   & 37.4  & 54.1  & \underline{34.4}  \\
    DISK~\cite{tyszkiewicz2020disk} + LG~\cite{lindenberger2023lightglue}    & 19.7   & 37.8  & 55.2  & \textbf{36.7}  \\
    MatchFormer-large~\cite{wang2022matchformer}   & \textbf{21.2}   & \textbf{39.6}  & \textbf{56.8}  & 32.4  \\
    \midrule
    MatchFormer-lite~\cite{wang2022matchformer}   & 19.8   & 36.9  & 54.0  & 32.1  \\
    + SCENES fine-tuning & \green{+0.1} & \green{+2.0} & \green{+2.2} & \redtext{-1.0}\\
     = Ours (MF) & 19.9  & \underline{38.9}  & \underline{56.2}  & 31.1 \\
    \midrule
    ASpanFormer~\cite{chen2022aspanformer} & 20.1   & 38.4  & 55.4  & 33.4  \\
    + SCENES fine-tuning  & \green{+1.1} & \green{+0.4} & \green{0.0} & \redtext{-0.4}\\
    = Ours (ASF) & \textbf{21.2}  & 38.8  &  55.4 & 33.0 \\ 
    \bottomrule
  \end{tabularx}
\end{table}

\subsection{Ablation study}

\begin{table}[!t]\centering\small
  \caption{Ablation study comparing variants of our design, evaluated with respect to relative pose estimation performance on the indoor EuRoC-MAV dataset \cite{burri2016eurocmav}. We report the area under the curve (AUC) at thresholds of $5\degree$, $10\degree$ and $20\degree$ and the matching precision (P) at a threshold of $5\cdot 10^{-4}$.
  The row ``w/o classification argmax'' denotes using the na\"ive epipolar classification loss, where $\mM^\textrm{epi}_{ij}=1$ for all pixels on the epipolar line, not just the highest probability pixel. The rows `perturb poses by $x\degree$' randomly perturb the ground truth translation and rotation by $x$ degrees.
  }
  \label{tab:ablation}
  \begin{tabularx}{\linewidth}{lCCCC}
    \toprule
    \multirow{2}{*}{Method} & \multicolumn{3}{c}{Pose Estimation AUC (\%)} & \multirow{2}{*}{P (\%)} \\
    \cmidrule(r){2-4} 
                        & $@5\degree$ & $@10\degree$ & $@20\degree$  &  \\
    \midrule
    MatchFormer-lite~\cite{wang2022matchformer}   & 3.0   & 11.6  & 25.1  & 35.0  \\
    Ours (MF) & 9.1  & \textbf{23.5}  &  \textbf{42.8} & 63.8  \\
    w/o fine-level loss & 5.5 & 15.2 & 29.4 & 61.2 \\
    w/o coarse-level loss & \textbf{10.0} & 20.7 & 34.4 & \textbf{69.9} \\
    w/o classification argmax & 0.0 & 0.0 & 0.0 & 0.0\\
    perturb poses by $1\degree$ & 8.4 & 20.2 & 36.5 & 63.1 \\
    perturb poses by $2\degree$ & 4.2 & 14.7 & 30.0 & 41.1 \\
    \bottomrule
  \end{tabularx}
\end{table}

In this section, we analyse the impact of design choices on our method. We evaluate modifications to the fine-tuning of the MatchFormer-lite model on the EuRoC-MAV dataset \cite{burri2016eurocmav}. The results of the ablations are in \cref{tab:ablation}, using the same metrics as before in \cref{tab:results_eurocmav}.
We evaluate the impact of three alterations to the full method: removing the fine-level loss ($\lambda=0$), removing the coarse-level loss ($\lambda=1$), and removing the argmax in \cref{eq:epipolar_classification}, setting $\mM^\textrm{epi}_{ij} = 1 \textrm{ iff } \bx_2^j \in e_1^i$ and zero otherwise. This sets the mask to 1 for all pixels on the epipolar line, not just the highest probability pixel.

If we remove the loss term for either the coarse or fine matching stage, the network focuses exclusively on the remaining task and its performance on the other deteriorates.
Removing the fine-level loss leads the network to create a large number of approximate matches, but these are too noisy to estimate accurate poses.
Removing the coarse-level loss leads the network to create fewer matches. These may be very accurate, but too few matches make the RANSAC pose estimator less robust and prone to fail entirely.
While this model does better for metrics where high accuracy is important, it does notably worse at the metrics with less stringent thresholds ($10\degree$ and $20\degree$).
On balance, a mixture of both losses leads to the overall most reliable performance.

Finally, as mentioned in \cref{subsec:epipolar_losses}, removing the argmax in the classification loss actively discourages the network to select good coarse matches.
The matching probability in the confidence matrix $\mC$ becomes spread-out along the entire epipolar line, which reduces the probability that an approximately correct coarse match will be selected.
Fine-tuning with this loss causes the model to diverge, failing to provide sufficient accurate matches for pose estimation.

\paragraph{Impact of Noisy Ground-truth Poses.}
We also evaluate the impact of noise in the ground truth poses on the effectiveness of the SCENES fine-tuning. We fine-tune the Matchformer-lite model on the EuRoC-MAV dataset as in \cref{sec:exp-euroc-mav}, but randomly perturb the ground truth poses (both rotation and translation) by magnitudes of one and two degrees. These results can also be found in \cref{tab:ablation}.
While seemingly small levels of pose noise affect the performance considerably, this is not too problematic.
As the results using our bootstrapping approach (cf. \cref{sec:exp-san-fran} and the supplementary material) show, existing keypoint matching methods already perform well enough to estimate fundamental matrices to sufficient accuracy for SCENES fine-tuning.

For further quantitative and qualitative results and 
ablations, we refer the interested reader to the appendix.

%% file: sec/5_conclusion.tex
\section{Discussion and Conclusion}

\paragraph{Limitations.}
One limitation is that the training signal is weaker than the fully-supervised approach, since a match predicted to be at the wrong place on the epipolar line does not incur a penalty.
Moreover, the epipolar classification loss in particular can be misleading, since it encourages the match to move towards the highest probability position on the epipolar line, which may be suboptimal and entrenches any biases learned by the pre-trained model.

\paragraph{Conclusion.}
We have proposed a method for estimating subpixel correspondences between two images, without requiring ground-truth correspondences for training.
To do so, we reformulate the standard classification and regression matching losses as epipolar losses, which only require ground-truth fundamental matrices during training.
These can be computed when the relative camera poses and intrinsics are known, 
or can be estimated by an adequate pre-trained matching network.
We develop two strategies for adapting pre-trained matching models to new domains: one where the fundamental matrices are provided, and another where they must be estimated, a bootstrapping approach.
We demonstrate compelling performance on challenging datasets, improving the pose estimation results of state-of-the-art methods using our weakly supervised and unsupervised finetuning strategies.
This opens the door to transfer learning or active learning applications.

\paragraph{Future work.}
Four directions warrant further investigation.
First, the bootstrapping approach could be made iterative, where the fundamental matrices are re-estimated every few epochs as the model improves.
This may reduce the impact of incorrect matches from the pre-trained network.
Second, the approach could be coupled with recent methods that infer camera poses given a sparse collection of images \cite{zhang2022relpose,lin2023relposepp}.
Third, given uncertainty estimates for the relative poses, the epipolar losses could be modified to reflect this uncertainty, transforming an epipolar line into a probability distribution on the image plane.
Finally, note that the true match must lie between the epipole (projection of the first camera centre into the second image) and the projection of the point at infinity.
This additional constraint could be used to improve the signal obtained from our epipolar losses.

\paragraph{Acknowledgements.} We are grateful for funding from EPSRC AIMS CDT EP/S024050/1 (D.K.), Continental AG (D.C.), and the Royal Academy of Engineering (RF/201819/18/163, J.H.). We would also like to thank Yash Bhalgat for valuable discussions.

%% file: sec/X_suppl.tex
\clearpage
\setcounter{page}{1}
\setcounter{section}{0}
\setcounter{table}{0}
\renewcommand\thesection{\Alph{section}}

\maketitlesupplementary

\section{Heuristic for Creating Image Pairs}

In this section, we describe in more detail the heuristics used to create image pairs for both training and testing, using only camera pose estimates. 

If ground-truth depth information is available, the overlap between two images can be computed and valid image pairs are those with an overlap within some range. This range is chosen to ensure there is enough overlap between the two images of a pair to compute accurate keypoints, but not so much overlap to make this task trivial.

Without ground-truth-depth, we assume a `pseudo-ground-truth-depth', and use this to compute approximate image overlaps. While this approximation is extremely rough, visual inspection of the generated image pairs showed that it nonetheless selects image pairs with the desired characteristic of enough, but not too much, overlap.

\paragraph{EuRoC-MAV.} 
For both the EuRoC-MAV training and testing datasets, we assume that each image was taken from inside a hemisphere. The hemisphere consists of a horizontal plane at a certain height $z_\textrm{plane} < z_\textrm{camera}$ and a spherical dome with radius $r_\textrm{sphere}$ centered at $(x_\textrm{camera}, y_\textrm{camera}, z_\textrm{plane})$.

For the machine room environments, we chose $z_\textrm{plane} = -2.0$m and $r_\textrm{sphere}= 10.0$m, and for the smaller room we chose  $z_\textrm{plane} = 0.0$m and $r_\textrm{sphere}= 3.0$m.

We also undistort the images in the EuRoC-MAV dataset and scale and crop them to a resolution of $640\times480$ pixels. 

\paragraph{San Francisco.}
The San Francisco training set images are provided together with GPS camera pose information. These are too inaccurate to use as supervision signal, but we are able to use them to generate image pairs. 

Since most images in the training set are taken from the middle of a road with multi-storey building on either side, we assume that each image was taken from inside a rectangular box aligned with the current driving direction. 
The left and right sides of the box are at a distance of $10$ meters from the camera, the bottom plane is $2$ meters below the camera, the front and back planes are at a distance of $25$ meters from the camera, and the top plane is at infinity.
We estimate the driving direction as the vector between locations of the preceding and the succeeding images.
Pixels whose rays intersect with the front or back planes are always counted as not overlapping with another image.

Since the test set contained comparatively few images, we selected appropriate image pairs manually. 

We scale the images so that their longest side has length $640$ pixels, then pad them to size $640\times640$ pixels.

\section{Further Ablation Study}
In this section, we give the results of some additional ablations (\cref{tab:ablation_eurocmav2,tab:ablation_sanfrancisco2}). 
Firstly, we investigate the impact of slight variations in the definition of $e^i_1 \in \mI_2$, the set of coordinates in the second image that form the epipolar line for the point $\bx_1^i \in \mI_1$. 
Since we are working with pixel patches of finite size, $e_1^i$ is the set of pixels that the epipolar line ``passes through''. Formally, if $w$ is the width of a single pixel patch, this means that
\begin{equation} \label{eq:theta}
    e^i_1 = \{\bx_2: \bx_2 \in \mI_2, d^\textrm{epi}(\bx^i_1, \bx_2) \leq \theta \frac{w}{2}\}
\end{equation}
where $\theta$ is a parameter determining how close the epipolar line has to pass to the center of the pixel patch to count as ``passing through'' the patch. In the experiments in the main paper, we set $\theta = \sqrt{2}$, but other values are possible. 
Secondly, we investigate the impact of adding or removing images from the original training set (ScanNet \cite{dai2017scannet} for indoor models, and Megadepth \cite{MegaDepthLi18} for outdoor models) to each batch, using the original correspondence losses for these samples.

\begin{table}[!t]\centering\small
  \caption{Ablation study comparing variants of our design, evaluated on the indoor EuRoC-MAV dataset \cite{burri2016eurocmav}. We report the area under the curve (AUC) at thresholds of $5\degree$, $10\degree$ and $20\degree$ and the matching precision (P) at a threshold of $5\cdot 10^{-4}$. 
  For a detailed explanation of the parameter $\theta$, please refer to the main body and in particular to \cref{eq:theta}.
  The row labelled `Updated version' refers to a finetuned model that was finetuned for longer than the original finetuned version (`Ours') but not as long as the ablations (26 epochs compared to 20 and 36 epochs respectively). 
  }
  \label{tab:ablation_eurocmav2} 
  \begin{tabularx}{\linewidth}{lCCCC}
    \toprule
    \multirow{2}{*}{Method} & \multicolumn{3}{c}{Pose Estimation AUC (\%)} & \multirow{2}{*}{P (\%)} \\
    \cmidrule(r){2-4} 
                        & $@5\degree$ & $@10\degree$ & $@20\degree$  &  \\
    \midrule
    MatchFormer-lite~\cite{wang2022matchformer}   & 3.0   & 11.6  & 25.1  & 35.0  \\
    Ours (MF) & 9.1  & 23.5 & 42.8 & 63.8  \\
    Updated version (MF) & 12.9 & 26.7 & 42.1 & 69.9 \\
    w/ $\theta = 1.0$ & 11.8 & 27.0 & 43.8 & \textbf{71.0} \\
    w/ $\theta = 3\sqrt{2}$ & \textbf{14.8} & \textbf{30.3} & \textbf{46.9} & 56.8 \\
    w/ ScanNet samples & 13.9 & 28.5 & 44.2 & 68.2 \\
    \bottomrule
  \end{tabularx}
\end{table}
\begin{table}[!t]\centering\small
  \caption{Ablation study comparing variants of our design, evaluated on the outdoor San Francisco dataset \cite{chen2011sanfrancisco} . We report the area under the curve (AUC) at thresholds of $5\degree$, $10\degree$ and $20\degree$ and the matching precision (P) at a threshold of $10^{-4}$. 
  For a detailed explanation of the parameter $\theta$, please refer to the main body and in particular to \cref{eq:theta}.
  }
  \label{tab:ablation_sanfrancisco2} 
  \begin{tabularx}{\linewidth}{lCCCC}
    \toprule
    \multirow{2}{*}{Method} & \multicolumn{3}{c}{Pose Estimation AUC (\%)} & \multirow{2}{*}{P (\%)} \\
    \cmidrule(r){2-4} 
                        & $@5\degree$ & $@10\degree$ & $@20\degree$  &  \\
    \midrule
    MatchFormer-lite~\cite{wang2022matchformer} & 19.8   & 36.9  & 54.0  & \textbf{32.1} \\
    Ours (MF) & \textbf{19.9}  & \textbf{38.9}  & \textbf{56.2}  & 31.1 \\
    w/ $\theta = 3\sqrt{2}$ & 19.7 & 38.1 & 55.2 & 31.4  \\
    w/o Megadepth samples & 18.3 & 35.6 & 52.7 & 30.1 \\
    \bottomrule
  \end{tabularx}
\end{table}

The additional ablations for the EuRoC-MAV dataset in \cref{tab:ablation_eurocmav2} are consistently stronger than the original finetuned model that we evaluated in the main paper. This suggests that the model whose results were reported in the main paper had not fully converged. This hypothesis that is supported by the similarly improved results for the `Updated version', which was trained for longer than the original finetuned version reported in the main paper, but not as long as the ablations in \cref{tab:ablation_eurocmav2} (26 epochs compared to 20 and 35 epochs respectively).

In any case, the consistently strong improvements with respect to the baseline model show the robustness of our finetuning method to moderate changes in certain parameters and the training algorithm.
We can note that a lower value for $\theta$ leads to lower pose estimation performance, as the pixel patch with maximum confidence on the epipolar line is more likely to be further from the correct match. The lower value for $\theta$ however leads to fewer coarse matches, so that the matches that are found are more likely to be accurate, increasing the precision.

The ablations for the San Francisco dataset in \cref{tab:ablation_sanfrancisco2} illustrate the importance of the added Megadepth samples to avoid excessively overfitting onto the San Francisco training set distribution, which is slightly different to the test set distribution. While increasing the parameter $\theta$ by a factor of $3$ slightly decreases the localisation performance, finetuning still improves performance.

\section{Additional Quantitative Results}
In this section we provide some additional quantitative results on the relative pose estimation task for the EuRoC-MAV and the San Francisco datasets. We report the fraction of relative pose estimates for which the angular rotation error and the angular translation error are below thresholds of $5\degree, 10\degree, 20\degree$, as well as the median angular rotation and translation errors in degrees (\cref{tab:results_eurocmav_2}, \cref{tab:results_sanfrancisco_2}).

\begin{table}[!t]
  \centering\footnotesize
  \caption{Relative pose estimation performance on the indoor EuRoC-MAV dataset \cite{burri2016eurocmav}. We report the fraction of angular rotation and translation errors below the thresholds $5\degree$, $10\degree$ and $20\degree$ in percent and the median angular rotation and translation errors in degree. Our SCENES method exhibits significant gains in performance across these metrics.}
  \label{tab:results_eurocmav_2}
  \begin{tabularx}{\linewidth}{lCCCC}
    \toprule
    \multirow{2}{*}{Method} & \multicolumn{4}{c}{Rotation} \\
    \cmidrule(r){2-5} 
                        & $<5\degree$ (\%) & $<10\degree$ (\%) & $<20\degree$ (\%)  & Median (\degree) \\
    \midrule
    SP~\cite{detone2018superpoint} + SG~\cite{sarlin2020superglue}      & 5.20   & 12.0   & 29.3   & 39.3  \\
    SP~\cite{detone2018superpoint} + LG~\cite{lindenberger2023lightglue}      & 54.0   & 67.6  & 77.1  & \underline{4.16}  \\
    DISK~\cite{tyszkiewicz2020disk} + LG~\cite{lindenberger2023lightglue}    & 44.1   & 56.5  & 64.7  & 6.86  \\
    MatchFormer-large~\cite{wang2022matchformer}   & 43.5   & 61.6  & 74.3  & 6.18  \\
    \midrule
    MatchFormer-lite~\cite{wang2022matchformer}   & 37.3   & 54.4  & 64.7  & 7.66  \\
    + SCENES finetuning    & \green{+17.6}  & \green{+26.0}  & \green{+29.6}  & \green{-3.40}  \\
                    = Ours (MF)        & \underline{54.9}          & \textbf{80.4} & \textbf{94.3}  &4.26  \\
    \midrule
    ASpanFormer~\cite{chen2022aspanformer} & 48.4   & 65.9  & 74.6  & 5.28  \\
    + SCENES finetuning   & \green{+10.3} & \green{+12.3} & \green{+15.9} & \green{-1.43} \\
    = Ours (ASF) & \textbf{58.7}  & \underline{78.2}  & \underline{90.5}  & \textbf{3.85}  \\ 
    \midrule \midrule
     & \multicolumn{4}{c}{Translation} \\
    \cmidrule(r){2-5} 
                        & $<5\degree$ (\%)& $<10\degree$ (\%) & $<20\degree$ (\%)  & Median (\degree) \\
    \midrule
    SP~\cite{detone2018superpoint} + SG~\cite{sarlin2020superglue}      & 2.80   & 7.53   & 17.2   & 48.6  \\
    SP~\cite{detone2018superpoint} + LG~\cite{lindenberger2023lightglue}      & 26.1   & \underline{50.9}  & 65.1  & \underline{9.74}  \\
    DISK~\cite{tyszkiewicz2020disk} + LG~\cite{lindenberger2023lightglue}    & 19.9   & 38.8  & 51.4  & 18.1  \\
    MatchFormer-large~\cite{wang2022matchformer}   & 18.8   & 38.8  & 56.2  & 14.0  \\
    \midrule
    MatchFormer-lite~\cite{wang2022matchformer}   & 11.8   & 30.3  & 48.5  & 21.5  \\
    + SCENES finetuning    & \green{+14.9}  & \green{+19.3}  & \green{+24.6}  & \green{-11.2}  \\
                    = Ours (MF)        & \underline{26.7}          & 49.6 & \textbf{73.1}  & 10.3  \\
    \midrule
    ASpanFormer~\cite{chen2022aspanformer} & 21.1   & 45.3  & 61.7  & 11.7  \\
    + SCENES finetuning   & \green{+7.9} & \green{+7.3} & \green{+10.7} & \green{-2.49} \\
    = Ours (ASF) & \textbf{29.0}  & \textbf{52.6}  & \underline{72.4}  & \textbf{9.21}  \\ 
    
    \bottomrule
  \end{tabularx}
\end{table}

\begin{table}[!t]
  \centering\footnotesize
  \caption{Relative pose estimation performance on the outdoor San Francisco dataset \cite{chen2011sanfrancisco}. We report the fraction of angular rotation and translation errors below the thresholds $5\degree$, $10\degree$ and $20\degree$ in percent and the median angular rotation and translation errors in degree.}
  \label{tab:results_sanfrancisco_2}
  \begin{tabularx}{\linewidth}{lCCCC}
    \toprule
    \multirow{2}{*}{Method} & \multicolumn{4}{c}{Rotation} \\
    \cmidrule(r){2-5} 
                        & $<5\degree$ (\%) & $<10\degree$ (\%) & $<20\degree$ (\%)  & Median (\degree) \\
    \midrule
    SP~\cite{detone2018superpoint} + SG~\cite{sarlin2020superglue}      & 7.95   & 17.4   & 32.7   & 33.0  \\
    SP~\cite{detone2018superpoint} + LG~\cite{lindenberger2023lightglue}      & 64.5   & 79.0  & 87.4  & 3.32  \\
    DISK~\cite{tyszkiewicz2020disk} + LG~\cite{lindenberger2023lightglue}    & 64.0   & \underline{80.8}  & \underline{89.0}  & 3.38  \\
    MatchFormer-large~\cite{wang2022matchformer}   & \textbf{65.8}   & \textbf{82.6}  & 88.7  & \underline{3.30}  \\
    \midrule
    MatchFormer-lite~\cite{wang2022matchformer}   & 63.4   & 79.0  & 87.2  & 3.41  \\
    + SCENES finetuning    & \green{+0.4}  & \green{+1.6}  & \green{+2.0}  & \redtext{+0.1}  \\
                    = Ours (MF)        & 63.8          & 80.6 & \textbf{89.2}  &3.42  \\
    \midrule
    ASpanFormer~\cite{chen2022aspanformer} & 63.4  & 79.7  & 87.2  & 3.47 \\
    + SCENES finetuning   & \green{+1.7} & \green{+0.2} & \green{+0.4} & \green{-0.18} \\
    = Ours (ASF) & \underline{65.1}  & 79.9  & 87.6  & \textbf{3.29}  \\ 
    \midrule \midrule
     & \multicolumn{4}{c}{Translation} \\
    \cmidrule(r){2-5} 
                        & $<5\degree$ (\%)& $<10\degree$ (\%) & $<20\degree$ (\%)  & Median (\degree) \\
    \midrule
    SP~\cite{detone2018superpoint} + SG~\cite{sarlin2020superglue}      & 10.6   & 26.3   & 49.9   & 20.0  \\
    SP~\cite{detone2018superpoint} + LG~\cite{lindenberger2023lightglue}      & 51.2   & 68.7  & 78.6  & 4.77  \\
    DISK~\cite{tyszkiewicz2020disk} + LG~\cite{lindenberger2023lightglue}    & 51.7   & 70.9  & 78.8  & 4.76  \\
    MatchFormer-large~\cite{wang2022matchformer}   & \underline{54.1}   & \textbf{71.5}  & \textbf{81.0}  & \textbf{4.30}  \\
    \midrule
    MatchFormer-lite~\cite{wang2022matchformer}   & 51.7   & 68.9  & 79.0  & 4.69  \\
    + SCENES finetuning    & \green{+2.8}  & \green{+2.4}  & \green{+1.4}  & \green{-0.35}  \\
                    = Ours (MF)        & \textbf{54.5}          & \underline{71.3} & 80.4  & \underline{4.34}  \\
    \midrule
    ASpanFormer~\cite{chen2022aspanformer} & 53.6   & \textbf{71.5}  & \underline{80.6}  & 4.42  \\
    + SCENES finetuning   & \redtext{-0.4} & \redtext{-0.6} & \redtext{-1.1} & \redtext{+0.12} \\
    = Ours (ASF) & 53.2  & 70.9  & 79.5  & 4.54  \\ 
    
    \bottomrule
  \end{tabularx}
\end{table}


\section{Additional Qualitative Results}
\begin{figure*}
    \centering
    \begin{subfigure}{\textwidth}
        \centering
        \includegraphics[width=0.45\linewidth]{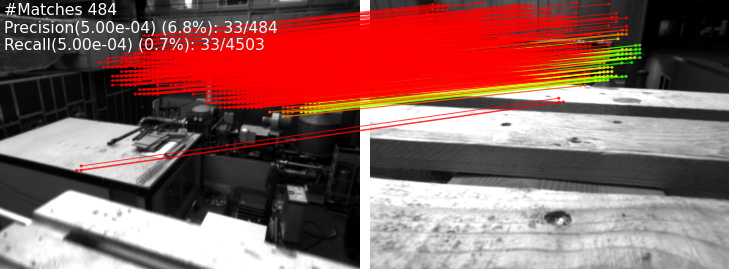}
        \includegraphics[width=0.45\linewidth]{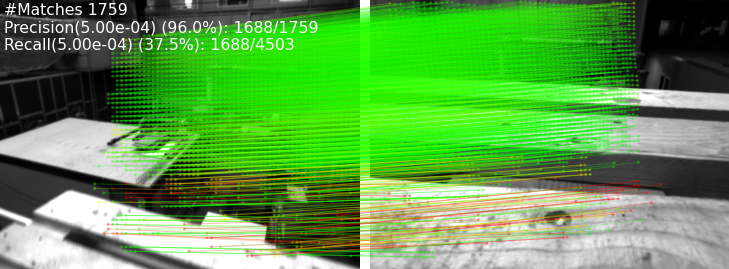}
        \caption{}
    \end{subfigure}
    \vspace{0.5cm} 
    \begin{subfigure}{\textwidth}
        \centering
        \includegraphics[width=0.45\linewidth]{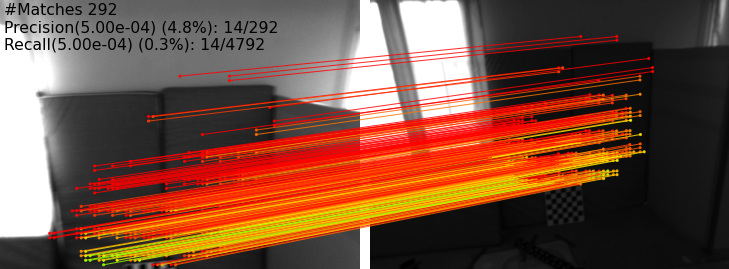}
        \includegraphics[width=0.45\linewidth]{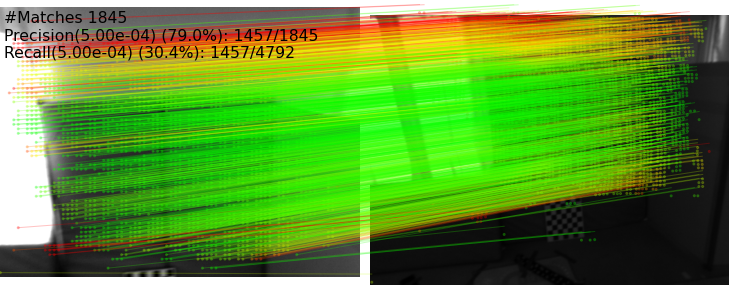}
        \caption{}
    \end{subfigure}
    \vspace{0.5cm} 
    \begin{subfigure}{\textwidth}
        \centering
        \includegraphics[width=0.45\linewidth]{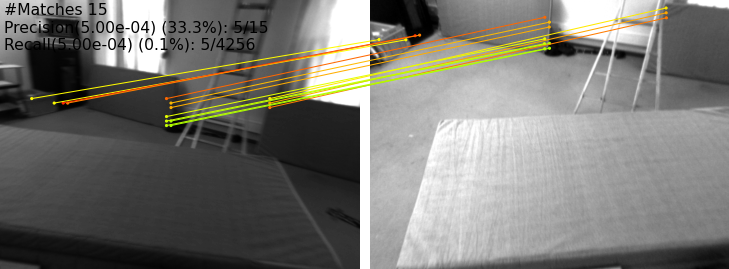}
        \includegraphics[width=0.45\linewidth]{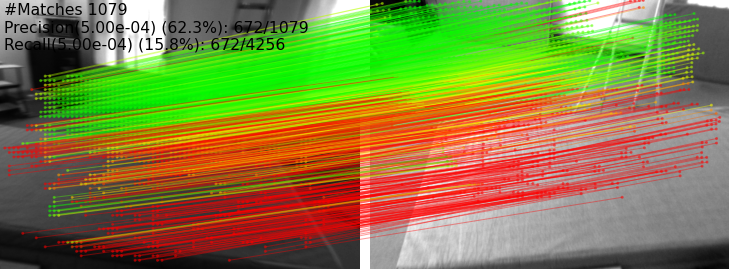}
        \caption{}
    \end{subfigure}
    \vspace{0.5cm} 
    \begin{subfigure}{\textwidth}
        \centering
        \includegraphics[width=0.45\linewidth]{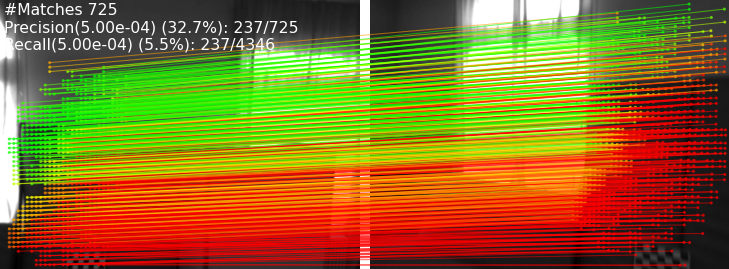}
        \includegraphics[width=0.45\linewidth]{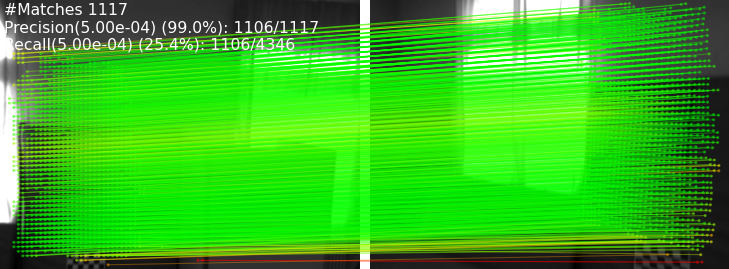}
        \caption{}
    \end{subfigure}
    \caption{Matches for the Matchformer-lite~\cite{wang2022matchformer} model before (left) and after (right) SCENES-finetuning on the EuRoC-MAV dataset.}
    \label{fig:qualitative_eurocmav_supp}
\end{figure*}

\begin{figure*}
    \centering
    \begin{subfigure}{\textwidth}
        \centering
        \includegraphics[width=0.45\linewidth]{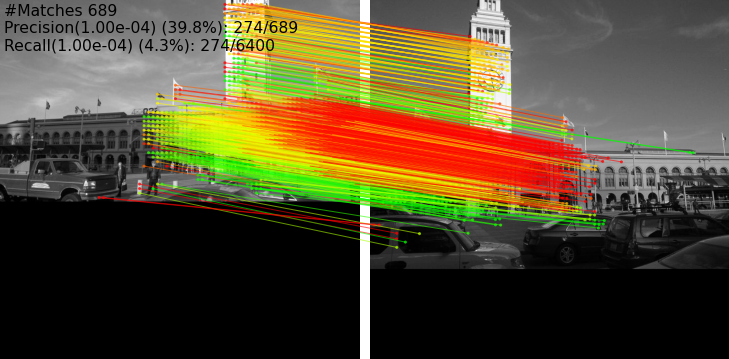}
        \includegraphics[width=0.45\linewidth]{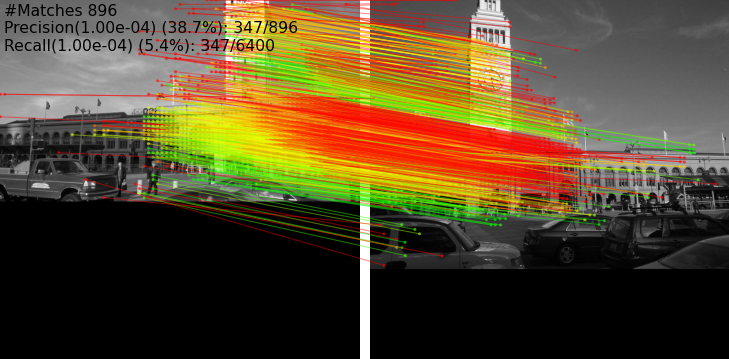}
        \caption{}
    \end{subfigure}
    \vspace{0.5cm} 
    \begin{subfigure}{\textwidth}
        \centering
        \includegraphics[width=0.45\linewidth]{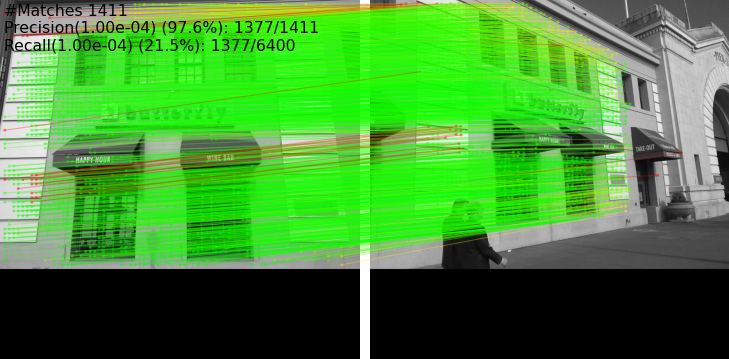}
        \includegraphics[width=0.45\linewidth]{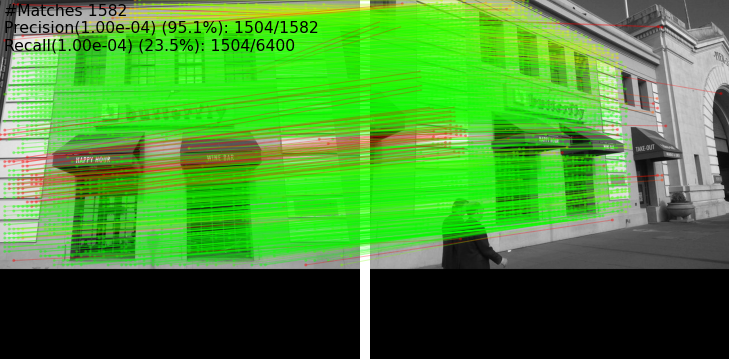}
        \caption{}
    \end{subfigure}
    \vspace{0.5cm} 
    \begin{subfigure}{\textwidth}
        \centering
        \includegraphics[width=0.45\linewidth]{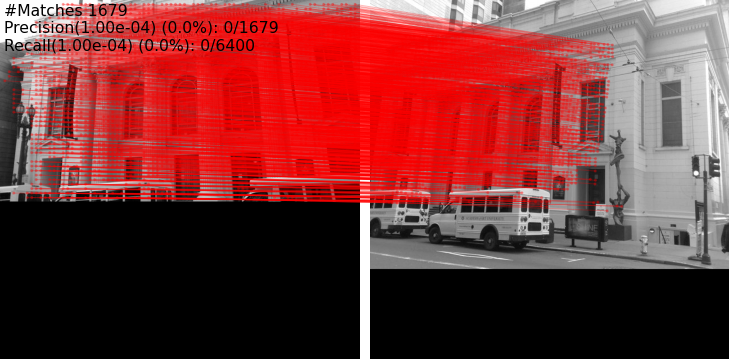}
        \includegraphics[width=0.45\linewidth]{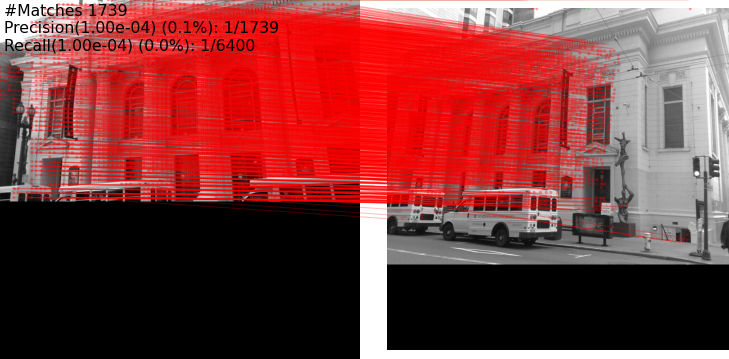}
        \caption{}
    \end{subfigure}
    \vspace{0.5cm} 
    \begin{subfigure}{\textwidth}
        \centering
        \includegraphics[width=0.45\linewidth]{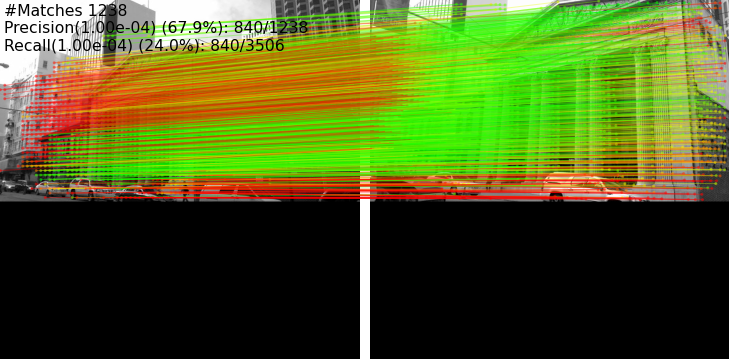}
        \includegraphics[width=0.45\linewidth]{figures/SanFrancisco_epipolarmatchformer_00336_fig00.png}
        \caption{}
    \end{subfigure}
    \caption{Matches for the Matchformer-lite~\cite{wang2022matchformer} model before (left) and after (right) SCENES-finetuning on the San Francisco dataset.}
    \label{fig:qualitative_sanfrancisco_supp}
\end{figure*}

In this section, we provide some additional qualitative results for finetuning the Matchformer-lite model on the EuRoC-MAV~\cite{burri2016eurocmav} dataset in \cref{fig:qualitative_eurocmav_supp} and the San Francisco~\cite{chen2011sanfrancisco} dataset in \cref{fig:qualitative_sanfrancisco_supp}. The lines indicating matched keypoints are coloured based on whether the matches' symmetric epipolar error is above or below a threshold of $5\cdot 10^{-4}$ for the indoor EuRoC-MAV dataset and $10^{-4}$ for the outdoor San Francisco dataset. 

The performance improvements on the EuRoc-MAV dataset after finetuning using ground-truth camera poses are very noticeable, with both more and more accurate keypoint matches being found.
The performance improvements on the San Francisco dataset after finetuning using our bootstrapping method without ground-truth poses are less stark but still evident in an increased number of accurate matches being found. The improvements are weakest for image pairs that the base model also fails on. This can be explained by the finetuning mostly increasing performance by improving the consistency of matches.

\section{Localisation on Aachen Day-Night}

We also provide results of models fine-tuned with our method as part of a larger image localisation pipeline. In particular, following previous papers \cite{sarlin2020superglue, sun2021loftr, chen2022aspanformer, lindenberger2023lightglue}, we use the HLoc pipeline \cite{sarlin2019coarse} and the Aachen Day-Night-v1.1 \cite{sattler2012aachen1, sattler2018aachen2} dataset. 

\paragraph{Implementation Details.} To create the image pairs for finetuning in the absence of even approximate camera pose information, we compute embeddings for each image using NetVLAD \cite{arandjelovic2016netvlad} and retain all image pairs for which these embeddings have a cosine similarity score of at least $0.24$. 

We finetune the Matchformer-lite outdoor model using the Aachen Day-Night training images using the same settings as for the San Francisco dataet, using our bootstrapped approach with fundamental matrices estimated by the base model, and adding an equal number of image pairs from the MegaDepth dataset \cite{MegaDepthLi18} to each batch. We also again set the weight decay to $0.01$, the learning rate to $1\cdot 10^{-4}$, and the batch size to $8$ image pairs taken from the Aachen Day-Night and MegaDepth datasets each.

\paragraph{Baselines.} We also provide results for the same set of keypoint localisation and matching algorithms as in our other experiments. To ensure the results are comparable, we resize all images so that their longest side has a length of $640$ pixels. To reduce the computational cost, we follow the pre-implemented algorithms in the HLoc pipeline and restrict the matches to return a maximum of one match per $8\times8$ pixel patch for the dense matching algorithms (Matchformer\cite{wang2022matchformer} and ASpanFormer\cite{chen2022aspanformer}).

\paragraph{Results.} The results as computed by the Long-Term Visual Localization benchmark \cite{toft2020visloc} server (\url{https://www.visuallocalization.net}) can be found in \cref{tab:aachen_results}.
Our bootstrapped SCENES method improves the localisation performance on the daytime images, but decreases the localisation performance on nighttime images. This is not entirely surprising -- the Aachen Day-Night training set consists only of daytime images, and finetuning processes generally sacrifice improved performance for slightly decreased generalisation ability.

 \begin{table}[!t]
  \centering\footnotesize
  \caption{Image Localisation performance on the Aachen Day-Night-v1.1 \cite{sattler2012aachen1,sattler2018aachen2} dataset using the HLoc\cite{sarlin2019coarse} toolbox.}
  \label{tab:aachen_results}
  \begin{tabularx}{\linewidth}{lCC}
    \toprule
    \multirow{2}{*}{Method} & Day & Night \\
    \cmidrule(r){2-3} 
                     & \multicolumn{2}{c}{$(0.25\textrm{m}, 2\degree)/(0.5\textrm{m}, 5\degree)/(1.0\textrm{m}, 10\degree)$} \\
    \midrule
    SP~\cite{detone2018superpoint} + SG~\cite{sarlin2020superglue}      & 87.1 / 94.3 / 98.3 &  69.1 / 88.0 / 97.9 \\
    SP~\cite{detone2018superpoint} + LG~\cite{lindenberger2023lightglue}      & 88.1 / 94.7 / 98.9 & 72.3 / 89.5 / 99.0 \\
    DISK~\cite{tyszkiewicz2020disk} + LG~\cite{lindenberger2023lightglue}    & \textbf{89.7} / \textbf{95.9} / \textbf{99.0} & \textbf{77.0} / \textbf{91.6} / \textbf{99.5} \\
    ASpanFormer~\cite{chen2022aspanformer} & 88.3 / 95.5 / 98.9 & 76.4 / 90.6 / 99.0\\
    MatchFormer-large~\cite{wang2022matchformer}   &87.7 / 95.8 / 98.9 &74.9 / 91.1 / 99.0\\
    \midrule
    MatchFormer-lite~\cite{wang2022matchformer}   &87.7 / 94.3 / 98.4 & 70.7 / 90.6 / 99.0 \\
    + SCENES finetuning    &\green{+0.2} / \green{+0.4} / \green{+0.1} & \redtext{-1.6} / \redtext{-0.5} / \redtext{-0.6}  \\
                    = Ours (MF)        & 87.9 / 94.7 / 98.5 & 69.1 / 90.1 / 98.4 \\

    \bottomrule
\end{tabularx}
\end{table}

\section{Bootstrapping on EuRoC-MAV}

In this section we provide the results of fine-tuning the MatchFormer-lite and ASpanFormer models on the EuRoC-MAV dataset without ground-truth pose supervision. Instead, we follow our bootstrapping strategy and compute the epipolar loss using the fundamental matrices estimated by the pre-trained models.

The dataset and evaluation are the same as in \cref{sec:exp-euroc-mav} in the main paper, and the results are available (together with the results from the main paper) in \cref{tab:results_eurocmav_bootstrap}.

\begin{table}[!t]
  \centering\small
  \caption{Relative pose estimation performance on the indoor EuRoC-MAV dataset \cite{burri2016eurocmav}. We report the area under the curve (AUC) at thresholds of $5\degree$, $10\degree$ and $20\degree$ and the matching precision (P) at a threshold of $5\cdot 10^{-4}$. Our SCENES method exhibits significant gains in performance across these metrics even when not using any ground truth poses for fine-tuning (`bootstrapping').}
  \label{tab:results_eurocmav_bootstrap}
  \begin{tabularx}{\linewidth}{lCCCC}
    \toprule
    \multirow{2}{*}{Method} & \multicolumn{3}{c}{Pose Estimation AUC (\%)} & \multirow{2}{*}{P (\%)} \\
    \cmidrule(r){2-4} 
                        & $@5\degree$ & $@10\degree$ & $@20\degree$  &  \\
    \midrule
    SP~\cite{detone2018superpoint} + SG~\cite{sarlin2020superglue}      & 0.6   & 2.0   & 4.4   & 17.3  \\
    SP~\cite{detone2018superpoint} + LG~\cite{lindenberger2023lightglue}      & 7.6   & 22.9  & 40.2  & 38.3  \\
    DISK~\cite{tyszkiewicz2020disk} + LG~\cite{lindenberger2023lightglue}    & 6.6   & 17.9  & 31.6  & 35.9  \\
    MatchFormer-large~\cite{wang2022matchformer}   & 5.1   & 16.0  & 31.8  & 38.4  \\
    \midrule
    MatchFormer-lite~\cite{wang2022matchformer}   & 3.0   & 11.6  & 25.1  & 35.0  \\
    + SCENES finetuning  & \green{+6.1}  & \green{+11.9}  & \green{+17.7}  & \green{+28.8}  \\
                    = Ours (MF)        & \underline{9.1}          & 23.5 & 42.8  & \underline{63.8}  \\
    \midrule
    MatchFormer-lite~\cite{wang2022matchformer}   & 3.0   & 11.6  & 25.1  & 35.0  \\
    + SCENES bootstrapping   & \green{+5.2}  & \green{+12.3}  & \green{+17.7}  & \green{+6.5}  \\
                    = Ours (MF-BS)        & 8.2          & 23.9 & 42.8  & 41.5  \\
    \midrule
    ASpanFormer~\cite{chen2022aspanformer} & 4.8   & 17.6  & 34.5  & 38.7  \\
    + SCENES finetuning   & \green{+6.8} & \green{+8.8} & \green{+10.2} & \green{+18.1} \\
    = Ours (ASF) & \textbf{11.6}  & \textbf{26.2}  & \textbf{44.7}  & \textbf{66.8}  \\ 
    \midrule
    ASpanFormer~\cite{chen2022aspanformer} & 4.8   & 17.6  & 34.5  & 38.7  \\
    + SCENES bootstrapping   & \green{+3.8} & \green{+7.1} & \green{+8.4} & \green{+3.4} \\
    = Ours (ASF-BS) & 8.6  & \underline{24.7}  & \underline{42.9}  & 42.1  \\ 
    \bottomrule
  \end{tabularx}
\end{table}

Even with the additional noise in the supervisory signal due to using more approximate estimated fundamental matrices, the performance gain due to the SCENES fine-tuning is almost as substantial as when using ground-truth relative poses and fundamental matrices.

We might explain this by noting that the epipolar loss already only provides an approximate supervisory signal that is least likely to be misleading when the detected pixel match is relatively accurate.
The estimated fundamental matrices used int he bootstrapping method are least accurate for image pairs that the pre-trained model performs worst on, but these are the image pairs that the epipolar loss is least useful even when using ground-truth fundamental matrices. The additional noise due to estimating fundamental matrices therefore does not do a lot of additional harm in those cases.
This hypothesis is supported by the decrease in performance of the bootstrapping method being largest for the smallest threshold; when the keypoint accuracy needs to be highest, the additional noise is most deleterious.

These results underscore that our bootstrapping method is a very viable approach to easily adapt a pre-trained pixel matching model to new domains.